\documentclass{article}
\usepackage{arxiv}

\pdfoutput=1
\usepackage[pdftex]{}

\usepackage{times}
\usepackage{epsfig}
\usepackage{graphicx}
\usepackage{amsmath}
\usepackage{amssymb}
\usepackage{amsthm}
\usepackage[table]{xcolor}
\usepackage{multirow}
\usepackage{booktabs}
\usepackage[accsupp]{axessibility}
\usepackage{bm}
\usepackage{xfrac}
\usepackage{makecell}
\usepackage{hyperref}
\usepackage{natbib}

\newcommand{\mypar}[1]{\vspace{3pt}
\noindent\textbf{#1~}}

\newcommand{\ppm}{\,\scriptsize$\pm$}
\newcommand{\zz}{\mathbf{z}}
\newcommand{\xx}{\mathbf{x}}

\newcommand{\eeps}{\boldsymbol{\epsilon}}
\newcommand{\dorowcolors}{\rowcolors{2}{gray!10}{white}}
\DeclareMathOperator{\EX}{\mathbb{E}}

\usepackage{graphicx,scalerel}
\newcommand\wh[1]{\hstretch{2}{\hat{\hstretch{.5}{#1}}}}
\newcommand\wt[1]{\hstretch{2}{\tilde{\hstretch{.5}{#1}}}}

\definecolor{myred}{rgb}{1, 0, 0}
\definecolor{mygreen}{rgb}{0, .67, 0}

\newcommand{\myFigA}[2]{
{\includegraphics[width=.225\linewidth]
{figures/ChrisFigs/sigma1_#1_sigma2_#2-1.png}}
}

\newcommand{\myFigB}[1]{
{\includegraphics[height=.383\linewidth]
{figures/ChrisFigs/#1.png}}
}

\title{NC-TTT: A Noise Contrastive Approach for Test-Time Training}

\date{} 					

\author{David Osowiechi\thanks{Equal contribution} \And Gustavo A. Vargas Hakim\footnotemark[1] \And Mehrdad Noori \And Milad Cheraghalikhani \And Ismail Ben Ayed \And Christian Desrosiers}
\date{LIVIA, ÉTS Montréal, Canada \\ International Laboratory on Learning Systems (ILLS), \\ McGILL - ETS - MILA - CNRS - Université Paris-Saclay - CentraleSupélec, Canada \texttt{gustavo-adolfo.vargas-hakim.1@ens.etsmtl.ca, david.osowiechi.1@ens.etsmtl.ca,
mehrdad.noori.1@ens.etsmtl.ca, milad.cheraghalikhani.1@ens.etsmtl.ca\\
ismail.benayed@etsmtl.ca,  christian.desrosiers@etsmtl.ca}}

\renewcommand{\shorttitle}{NC-TTT: A Noise Contrastive Approach for Test-Time Training}

\hypersetup{
pdftitle={main},
pdfsubject={cs.ML, cs.AI, cs.CV},
pdfauthor={Gustavo A. Vargas Hakim, David Osowiechi, Mehrdad Noori, Milad Cheraghalikhani, Ismail Ben Ayed, Christian Desrosiers},
pdfkeywords={Test-time Adaptation, Image Classification, Mutual Information, Unsupervised Training},
}

\begin{document}
\maketitle

\begin{abstract}
   Despite their exceptional performance in vision tasks, deep learning models often struggle when faced with domain shifts during testing. Test-Time Training (TTT) methods have recently gained popularity by their ability to enhance the robustness of models through the addition of an auxiliary objective that is jointly optimized with the main task. Being strictly unsupervised, this auxiliary objective is used at test time to adapt the model without any access to labels. In this work, we propose Noise-Contrastive Test-Time Training (NC-TTT), a novel unsupervised TTT technique based on the discrimination of noisy feature maps. By learning to classify noisy views of projected feature maps, and then adapting the model accordingly on new domains, classification performance can be recovered by an important margin. Experiments on several popular test-time adaptation baselines demonstrate the advantages of our method compared to recent approaches for this task. The code can be found at: \url{https://github.com/GustavoVargasHakim/NCTTT.git}
\end{abstract}

\section{Introduction}
\label{sec:intro}

A crucial requirement for the success of traditional deep learning methods is that training and testing data should be sampled from the same distribution. As widely shown in the literature \cite{Recht2018,visda}, this assumption rarely holds in practice, and a model's performance can drop dramatically in the presence of domain shifts. The field of Domain Adaptation (DA) has emerged to address this important issue, proposing various mechanisms that adapt learning algorithms to new domains. 

In the realm of domain adaptation, two notable directions of research have surfaced: Domain Generalization and Test-Time Adaptation. Domain Generalization (DG) approaches \cite{dg1,dg2,dg3,dg4,dgsurvey} typically train a model with an extensive source dataset encompassing diverse domains and augmentations, so that it can achieve a good performance on test examples from unseen domains, without retraining. 

Conversely, Test-Time Adaptation (TTA) \cite{tent2021,sita2021,lame2022} entails the dynamic adjustment of the model to test data in real-time, typically adapting to subsets of the new domain, such as mini-batches. TTA presents a challenging, yet practical problem as it functions without supervision for test samples or access to the source domain data. While they do not require training data from diverse domains as DG approaches, TTA methods are often susceptible to the choice of unsupervised loss used at test time, a factor that can substantially influence their overall performance.
Test-Time Training (TTT), as presented in \cite{ttt,ttt++,tttmask,tttflow, clust3}, offers a compelling alternative to TTA. In TTT, an auxiliary task is learned from the training data (source domain) and subsequently applied during test-time to refine the model. Generally, unsupervised and self-supervised tasks are selected for their capacity to support an adaptable process, without relying on labeled data. Finally, employing a dual-task training approach in the source domain allows the model to be more confident at test time, as it is already familiar with the auxiliary loss.

Motivated by recent developments in machine learning using Noise-Contrastive Estimation (NCE) \cite{mnih2013learning,oord2018representation,aneja2021contrastive}, we introduce a Noise-Contrastive Test-Time-Training (NC-TTT) method that efficiently learns the distribution of sources samples by contrasting it with a noisy distribution. This is achieved by training a discriminator that learns to distinguish noisy out-of-distribution (OOD) features from in-distribution ones. At test time, the output of the discriminator is used to guide the adaptation process, modifying the parameters of the network encoder so that it produces features that match in-distribution ones.

Our contributions can be summarized as follows:
\begin{itemize}
    \item We present an innovative Test-Time Training approach inspired by the paradigm of Noise-Constrastive Estimation (NCE). While NCE was initially proposed for generative models as a way to learn a data distribution without having to explicitly compute the partition function \cite{gutmann2010noise,mnih2013learning}, and later employed for unsupervised representation learning \cite{aneja2021contrastive,oord2018representation}, our work is the first to show the usefulness of this paradigm for test-time training.
    \item We motivate our method with a principled and efficient framework deriving from density estimation, and use this framework to guide the selection of important hyperparameters.     
    \item In a comprehensive set of experiments, we expose our NC-TTT method to a variety of challenging TTA scenarios, each featuring unique types of domain shifts. Results of these experiments demonstrate the superior performance of our method compared to recent approaches for this problem.
\end{itemize}
The subsequent sections of this paper are structured as follows. Section~\ref{sec:related} reviews prior research on TTA, TTT, and NCE. Section~\ref{sec:method} presents our NC-TTT method along with the experimental framework for its evaluation, detailed in Section~\ref{sec:experiments}. Section~\ref{sec:results} offers experimental results and discussions, while Section~\ref{sec:conclusions} concludes the paper with final remarks.

\section{Related work}
\label{sec:related}

\mypar{Test-Time Adaptation.} TTA is the challenging problem of adapting a pre-trained model from a source domain to an unlabeled target domain in an online manner (i.e., on a batch-wise basis). In this problem, it is assumed that the model no longer has access to source samples, making the setting more realistic and applicable as an \emph{off-the-shelf} tool. Finally, the online nature of TTA also limits the possibility of computing accurate target data distributions, specially when the number of samples is low. 

Two classic TTA methods have prevailed in the literature, Prediction Time Batch Normalization (PTBN) \cite{PTBN} and Test-Time Adaptation by Entropy Minimization (TENT) \cite{tent2021}. The former consists in simply recomputing the statistics from each batch of data inside the batch norm layers, instead of using the frozen source statistics. The later goes one step further by minimizing the entropy loss on the model's predictions and updating only the affine parameters of the batch norm layers. Recently, LAME \cite{lame2022} introduced a closed-form optimization mechanism that acts on the model's predictions for target images. This method is based on the Laplacian of the feature maps, which enforces their clustering based on similarity. A more detailed presentation of TTA approaches can be found in \cite{liang2023comprehensive}. 

\mypar{Test-Time Training.} TTA methods assume the existence of an implicit property in the model that can be linked to accuracy and can be used for adaptation at test time (e.g., entropy \cite{tent2021}). In contrast, TTT techniques explicitly introduce a given property by learning a secondary task alongside the main classification task at training. As seminal work in the field, TTT \cite{ttt} introduced a Y-shaped architecture allowing for a self-supervised rotation prediction task. This sub-network can be attached to any layer of a CNN. Formally, the overall TTT objective is composed of a supervised loss $\mathcal{L}_{sup}$ (e.g., cross-entropy) and an auxiliary, task-dependent loss $\mathcal{L}_{aux}$, as follows:
\begin{equation}
    \mathcal{L}_{TTT} = \mathcal{L}_{sup} + \lambda \mathcal{L}_{aux}
    \label{eq:ttt}
\end{equation}
The auxiliary loss is used at test time to update the model's encoder, reconditioning the features into being more similar to those from the source domain. TTT++ \cite{ttt++} proposed using contrastive learning as the secondary task, while also preserving statistical information from the source domain's feature maps to align the test-time features. Similarly, TTT-MAE \cite{mae} used Masked Autoencoder (MAE) \cite{tttmask} image reconstruction as the auxiliary task. Normalizing Flows (NF) \cite{realnvp,glow} have also been employed in TTTFlow \cite{tttflow}, adapting the feature encoder at test time by approximating a likelihood-based domain shift detector. Unlike previous approaches, TTTFlow requires two separate training procedures for the original model and the NF network, which makes source training more complex. Recently, ClusT3 \cite{clust3} introduced an unsupervised secondary task where the projected features of a given layer are clustered using a mutual information maximization objective. Although ClusT3 achieves competitive results, the hyperparameters of this method (e.g., number clusters) are dataset dependent, which limits its generalization capabilities. 

\mypar{Noise-contrastive estimation (NCE).} Our work is also related to NCE, a useful tool to model unknown distributions by \emph{comparison} \cite{gutmann2010noise}. In NCE, a dataset is contrasted against a set of noisy points drawn by an arbitrary distribution. A discriminator is then trained to distinguish between both sets, thereby learning the original dataset's properties. This approach has been employed to learn word embeddings \cite{mnih2013learning}, training Variational Autoencoders \cite{aneja2021contrastive}, and self-supervised learning (InfoNCE) \cite{oord2018representation}, among others. To our knowledge, this work is the first to  investigate the potential of NCE for test-time training. We hypothesize that NCE is well suited to estimate the source domain distribution at training time, and that this estimation can be used in an unsupervised manner at test time to adapt a model to target domain samples.

\begin{figure*}[ht!]
    \centering
\includegraphics[width=.85\linewidth]
{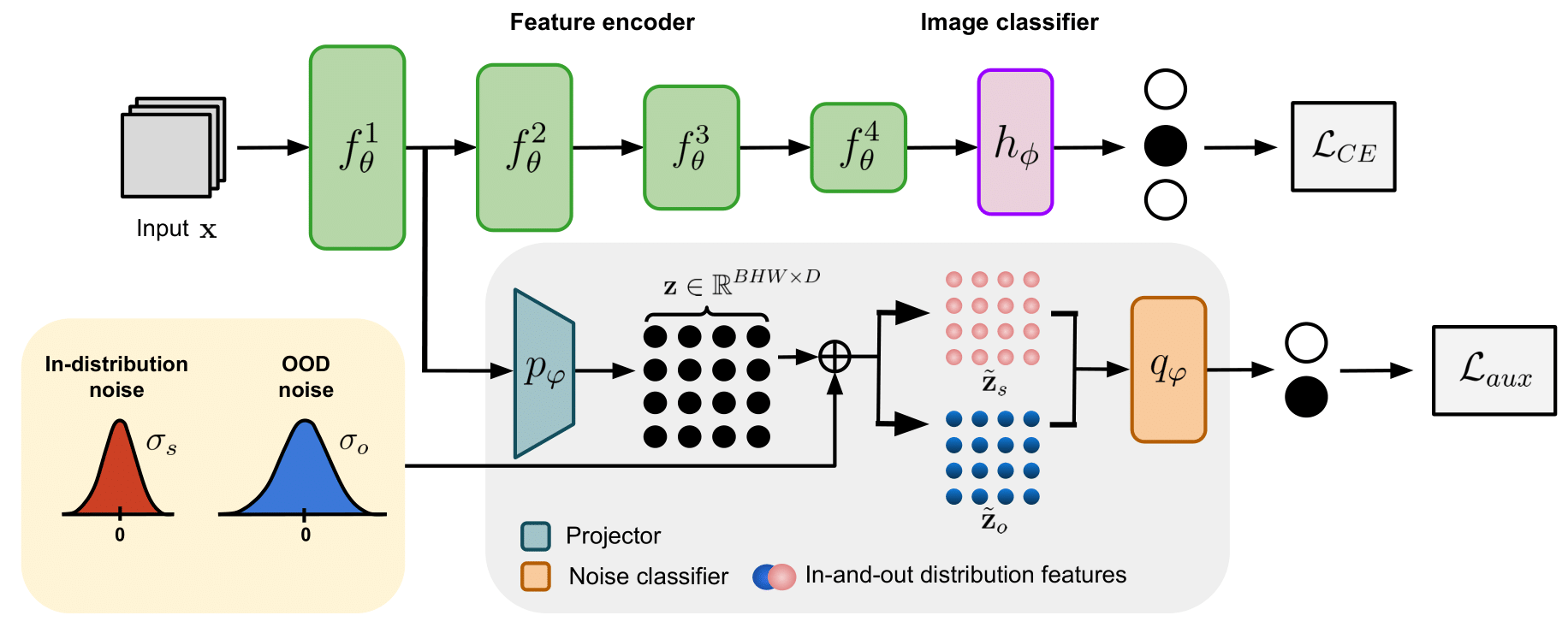}
\vspace*{-1mm}
    \caption{Overview of our Noise-Contrastive Test-Time-Training (NC-TTT) method. The auxiliary module comprises a linear projector $p_{\varphi}$ that reduces the scale of features, and a classifier $q_{\varphi}$ to discriminate between two different noisy views of the reduced features.
    \vspace*{-4mm}}
    \label{fig:noisenet}
\end{figure*}

\section{Methodology}
\label{sec:method}

We begin by presenting an overview of our NC-TTT method for Test-Time Training. We then proceed to detail the Noise-Contrastive Estimation framework on which it is grounded.

\subsection{The proposed method}

The problem of Test-Time Training can be formally defined as follows. Let the source domain be represented by a joint distribution $\mathcal{P}(\mathcal{X}_{s}, \mathcal{Y}_{s})$ , where $\mathcal{X}_{s}$ and $\mathcal{Y}_{s}$ correspond to the image and labels spaces, respectively. Likewise, denote as $\mathcal{P}(\mathcal{X}_{t}, \mathcal{Y}_{t})$ the target domain distribution, with $\mathcal{X}_{t}$ and $\mathcal{Y}_{t}$ as the respective target images and labels. Following previous research, we consider the likelihood shift \cite{lame2022} between source and target datasets, expressed as $\mathcal{P}(\mathcal{X}_{s}|\mathcal{Y}_{s}) \neq \mathcal{P}(\mathcal{X}_{t}|\mathcal{Y}_{t})$, and assume the label space to be the same between domains ($\mathcal{Y}_{s} = \mathcal{Y}_{t}$). Given a model $F: \mathcal{X} \rightarrow \mathcal{Y}$ trained on source data $(\xx,y) \in \mathcal{X}_{s}\times\mathcal{Y}_{s} $, the goal of TTT is to adapt this model to target domain examples from $\mathcal{X}_t$ at test time, without having access to source samples or target labels.   

As shown in Fig.~\ref{fig:noisenet}, our NC-TTT model follows the same Y-shaped architecture as in previous works, with the first branch corresponding to the main classification task and the second one to the auxiliary TTT task. The classification branch can be defined as $F_{\theta,\phi} = (h_{\phi} \circ f_\theta)$ where $f_{\theta} = (f^L_{\theta} \circ \ldots \circ f^1_{\theta})$ is an encoder that transforms images into feature maps via $L$ convolutional layers (blocks) and $h_{\phi}$ is a classification head that takes features from the last encoder layer and outputs the class probabilities. This branch is trained with a standard cross-entropy loss $\mathcal{L}_{CE}$

Following recent TTT approaches \cite{tttflow, clust3}, our auxiliary task operates on the features of the encoder. Without loss of generality, we suppose that the features come from layer $\ell$ of the encoder and denote as $f^\ell_{\theta}(\xx) \in \mathbb{R}^{B \times W \times H \times D}$ the $D$ feature maps of size $W\!\times\!H$ for a batch of $B$ images. We first reshape these feature maps to a $(BW\!H)\!\times\!D$ feature matrix and then use a linear projector to reduce its dimensionality, giving projected features $\zz = p_{\varphi}(f^\ell_{\theta}(\xx)) \in \mathbb{R}^{BW\!H \times d}$ with $d \ll D$. Next, we generate two noisy versions of $\zz$, an in-distribution version $\wt{\zz}_s = \zz + \eeps_s$, $\eeps_s \sim \mathcal{N}(\bm{0}, \sigma_s^2 I)$, and an out-of-distribution (OOD) version $\wt{\zz}_o = \zz + \eeps_o$, $\eeps_o \sim \mathcal{N}(\bm{0}, \sigma_o^2 I)$ where $\sigma_o > \sigma_s$. These noisy features are fed into a discriminator $q_\varphi$ which predicts in-distribution probabilities $[0,1]^{BW\!H}$. This discriminator, which is built using two linear layers with ReLU in between, is trained by minimizing loss $\mathcal{L}_{aux}$ computing the binary cross-entropy between the predicted probabilities and \emph{soft-labels} which will be described in the next section. To update the encoder parameters at test-time, as we do not have class labels, we only compute gradients from $\mathcal{L}_{aux}$. 

\begin{figure*}[ht!]
    \centering\setlength{\tabcolsep}{2pt}
    \begin{tabular}{llll}
    \myFigA{point01}{1} & \myFigA{point01}{5} & \myFigA{point01}{10} & \myFigA{point01}{50} \\[4pt]
    \myFigA{point05}{1} & \myFigA{point05}{5} & \myFigA{point05}{10} & \myFigA{point05}{50} \\
    \end{tabular}
    \vspace*{-1mm}
    \caption{Posterior probability $p(y_s = 1|\zz)$ of 2D points with different pairs $(\sigma_s, \sigma_o)$. The in-domain \emph{influence} expands by increasing $\sigma_o$ for a fixed $\sigma_s$ (see difference row-wise). Furthermore, this region is more regular when $\sigma_s$ increases when $\sigma_o$ is fixed (see difference column-wise).
    \vspace*{-4mm}}
    \label{fig:sigma1_sigma2}
\end{figure*}

\begin{figure}[ht!]
    \centering\setlength{\tabcolsep}{4pt}
    \begin{tabular}{ll}
    \myFigB{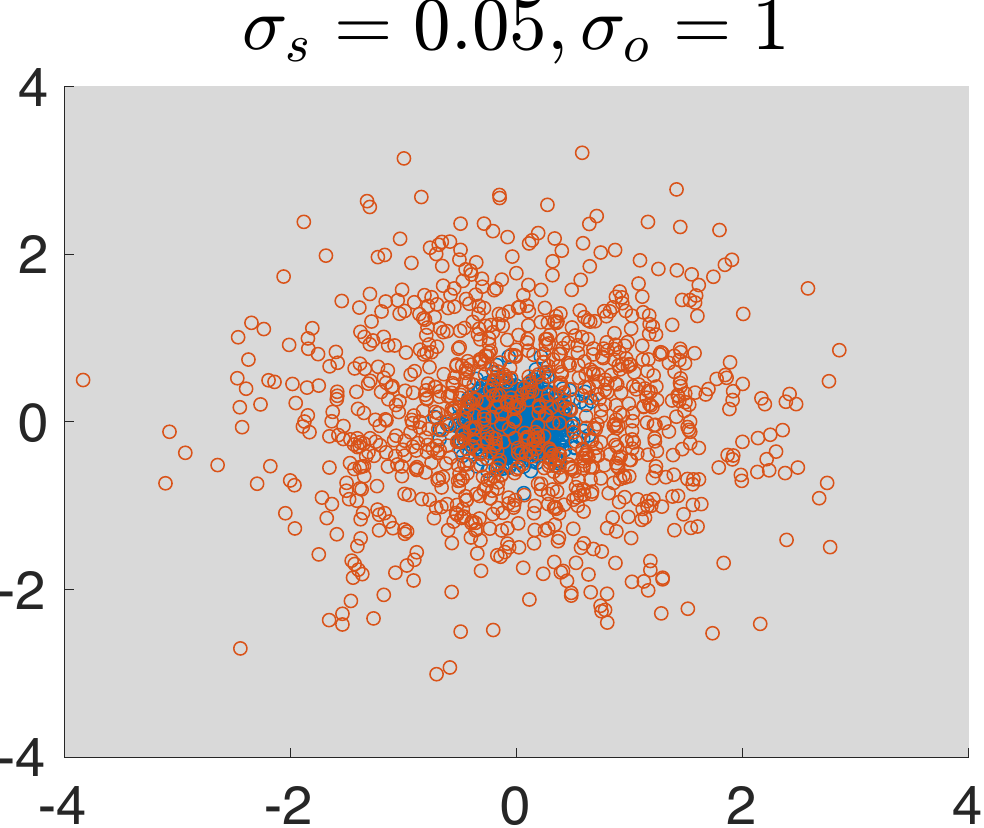} &
    \myFigB{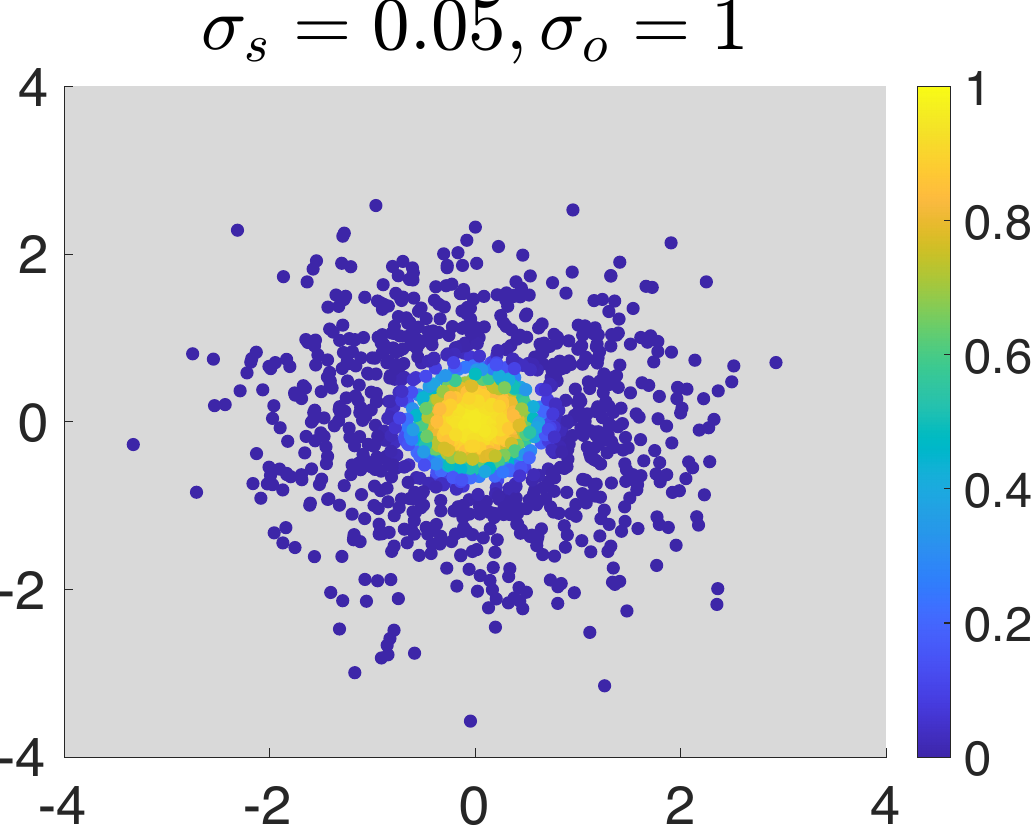} 
    \end{tabular}

    \vspace*{-1mm}
    \caption{Noise 2D vectors sampled with $\sigma_{s} = 0.05$ and $\sigma_{o} = 1$ (\emph{left}). The overlapping of both distributions can be overcome by assigning a probability to each point based on our threshold method.
    \vspace*{-4mm}}
    \label{fig:samples_probs}
\end{figure}

\subsection{Noise-contrastive Test-time Training}

We now present our noise-contrastive strategy for test-time training. Let us denote as $p_s(\zz)$ the probability of features from the source domain. Our method employs a density estimation strategy to learn $p_s(\zz)$ from training source examples $\mathcal{D}_s = \{\zz_i\}_{i=1}^{N_s}$, where $N_s=BW\!H$. Afterwards, it uses the estimated distribution $\wh{p}_s(\zz)$ to adapt the model to distribution shifts at test time.

\mypar{Estimating the source distribution.} We consider the well-known kernel density estimation approach to model $p_s(\zz)$. This approach puts a small probability mass around each training example $\xx_i \in \mathcal{D}_s$, in the shape of a $D$-dimensional Gaussian with isotropic variance $\Sigma_s = \sigma_s^2 I$, and then estimates the distribution as
\begin{equation}\label{eq:density_estimation}
\wh{p}_s(\zz) \, = \, \frac{1}{N_s(2\pi\big)^{D/2}\sigma_s^D}\sum_{i=1}^{N_s} \exp\!\left(\!-\frac{1}{2\sigma_s^2}\|\zz - \zz_i\|^2\right)
\end{equation}
At test-time, one could use this probability estimation to define an adaption objective $\mathcal{L}_{aux}$ that minimizes the negative log-likelihood of test examples $\mathcal{D}_t = \{\zz_j\}_{j=1}^{N_t}$:
\begin{equation}
\mathcal{L}_{aux} \, = \, -\frac{1}{N_t}\sum_{j=1}^{N_t}\log \,\wh{p}_s(\zz_j).
\end{equation}
However, this simple approach faces two important issues. First, estimating the density in high-dimensional space is problematic since moving away from a training example quickly reduces the probability to zero. Second, the training examples from the source domain are no longer available at test time, hence the density of samples in Eq. (\ref{eq:density_estimation}) cannot be evaluated.

To overcome these issues, we propose a \emph{noise contrastive} approach, which uses a discriminator to learn feature distribution $p_s(\zz)$. Toward this goal, we contrast $p_s(\zz)$ with an out-of-domain  distribution $p_o(\zz)$ which is also estimated using Eq. (\ref{eq:density_estimation}) but replacing the variance with $\sigma_o^2$, where $\sigma_o > \sigma_s$. Let $y_s$ be a domain indicator variable such that $y_s=1$ if an example is from the source domain, else $y_s=0$. Assuming equal priors $p(y_s=1) = p(y_s=0)$, we can use Bayes' theorem to get the posterior
\begin{equation}
p(y_s=1 \, | \, \zz) \, = \, \frac{\wh{p}_s(\zz)}{\wh{p}_s(\zz) \, + \, \wh{p}_o(\zz)}.
\end{equation}
To illustrate this model, we show in Figure \ref{fig:sigma1_sigma2} the probability $p(y_s=1 \, | \, \zz)$ obtained for different values of $\sigma_s$ and $\sigma_o$, when training with randomly-sampled 2D points. For a fixed $\sigma_s$, increasing $\sigma_o$ expands the in-domain region around the training samples. Likewise, for the same $\sigma_o$, using a greater $\sigma_s$ gives a larger and more regular (less determined by individual points) in-domain region. 

\mypar{Training the disciminator.} To train the discriminator $q_{\varphi}(\cdot)$, for each training example $\zz_i \!\in\! \mathcal{D}_s$, we generate $2M$ samples $\wt{\zz}_{i,m} \!=\! \zz_i + \eeps_{i,m}$, the first $M$ from the in-domain distribution, i.e. $\eeps_{i,m} \sim \mathcal{N}(\bm{0}, \sigma_s^2 I)$, and the other $M$ ones from the noisier out-of-domain distribution, i.e. $\eeps_{i,m} \sim \mathcal{N}(\bm{0}, \sigma_o^2 I)$. For these samples, we assume that $\exp(-\| \wt{\zz}_{i,m} - \zz_j \|_2^2/2\sigma_s^2) \approx 0$, for $j \neq i$, hence the posterior simplifies to
\begin{equation}\label{eq:posterior}
\begin{split}
& p(y_s=1 \, | \, \wt{\zz}_{i,m}) \, = \, \\[-6pt]
& \ \qquad
\frac{
\sigma_s^{-D} \exp\!\left(\!-\frac{1}{2\sigma_s^2}\|\eeps_{i,m}\|^2\right)
}{
\sigma_s^{-D} \exp\!\left(\!-\frac{1}{2\sigma_s^2}\|\eeps_{i,m}\|^2\right) + \sigma_o^{-D} \exp\!\left(\!-\frac{1}{2\sigma_o^2}\|\eeps_{i,m}\|^2\right)
}
\end{split}
\end{equation}
where $\eeps_{i,m} = \wt{\zz}_{i,m}-\zz_i$. For large values of $D$, this formulation is numerically unstable it leads to \emph{division by zero} errors. Instead, we use an equivalent formulation $p(y_s=1 \, | \, \wt{\zz}) = \mathrm{sigmoid}(u)$, where pre-activation ``logit'' $u$ is given by
\begin{equation}\label{eq:logit}
u \, = \, \frac{1}{2}\left(\frac{1}{\sigma_o^2} - \frac{1}{\sigma_s^2} \right) \|\eeps_{i,m}\|^2 \, + \, D\log\left(\frac{\sigma_o}{\sigma_s}\right)
\end{equation}
See Appendix A in the supplementary material for a proof. The in-domain region, $p(y_s=1 \, | \, \wt{\zz}) \geq 0.5$, which corresponds to the case where $u\geq 0$, is thus defined by the following condition:
\begin{equation}
\|\eeps_{i,m}\| \, \leq \, \sigma_s \sigma_o \sqrt{\frac{2D}{(\sigma_s^2 - \sigma_o^2)}\log\left(\frac{\sigma_s}{\sigma_o}\right)}
\end{equation}
Figure \ref{fig:samples_probs} shows examples of noise vectors $\eeps$ sampled with $\sigma_s\!=\!0.05$ and $\sigma_o\!=\!1$ (\emph{left}), and their corresponding posterior probability (\emph{right}). As can be seen, the posterior probability correctly separates in-distribution samples from OOD ones. Doing so, it overcomes the problem of having OOD samples that are similar to in-distribution ones (red circles near the center), which would confuse the discriminator during training. 

Using these samples $\wt{\zz}_{i,m}$, we train the discriminator $q_{\varphi}(\cdot)$ by minimizing the cross-entropy between its prediction and the soft-label $\wt{p}_{i,m} = p(y_s=1 \, | \, \wt{\zz}_{i,m})$:
\begin{equation}
\begin{split}
\mathcal{L}_{aux} \, = \, &
- \frac{1}{2MN_s} \sum_{i=1}^{N_s} \sum_{m=1}^{2M} \wt{p}_{i,m} \log q_{\varphi}(\wt{\zz}_{i,m})\\
& \quad+ \big(1-\wt{p}_{i,m}\big)\log\big(1-q_{\varphi}(\wt{\zz}_{i,m})\big)
\end{split}
\label{eq:aux_loss}
\end{equation}

\mypar{Adapting the model at test time.} During inference, we adapt the parameters of the encoder in layers where the auxiliary loss is computed, as well as those of preceding layers. The adaptation modifies the encoder so that the trained discriminator $q_{\varphi}(\cdot)$ perceives the encoded features $\{\zz_j\}_{j=1}^{N_t}$ of test examples as being in-distribution. This is achieved by minimizing the following test-time loss:  
\begin{equation}
\mathcal{L}_{aux}^{test} \, = \,
- \frac{1}{N_t} \sum_{j=1}^{N_t} \log q_{\varphi}(\zz_{j})
\end{equation}

\begin{figure}
    \centering
    \includegraphics[height=.42\linewidth]
    {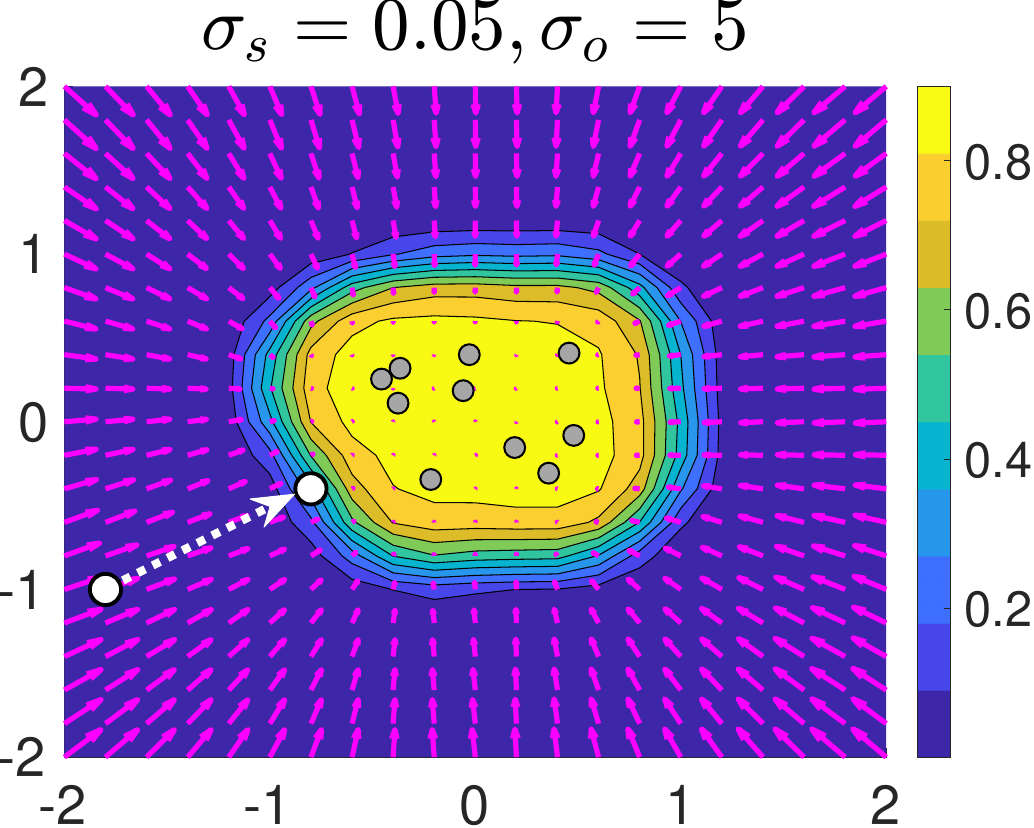}
    \vspace*{-1mm}
    \caption{Heatmap of in-distribution probabilities, i.e., $p(y_s\!=\!1 \, | \, \zz)$ approximated by $q_{\varphi}(\zz)$ in our model, and spatial gradient of log-likelihood function, i.e. $\nabla \log q_{\varphi}(\zz)$, which is used as test-time adaptation objective. The arrow shows how an OOD test sample (white point) is adapted toward the source distribution.
    \vspace*{-4mm}
    }
    \label{fig:gradient}
\end{figure}

As illustrated in Fig.~\ref{fig:gradient}, our method models the in-distribution probability $p(y_s\!=\!1 \, | \, \zz)$ using NCE and then approximates this distribution with discriminator $q_{\varphi}(\cdot)$. At test time, the encoder is updated to move OOD features (white point) toward the source distribution, making them more suitable for the source-trained classifier. Thanks to the non-zero in-distribution noise ($\sigma_s\!>\! 0$), we avoid over-adapting the encoder (the white point stops at the border of the in-distribution region and not at a training sample), a problem often found in other TTT approaches.

\subsection{Selecting the distribution variances}

\begin{figure}[ht!]
\centering
    {\includegraphics[width=.6\linewidth]
{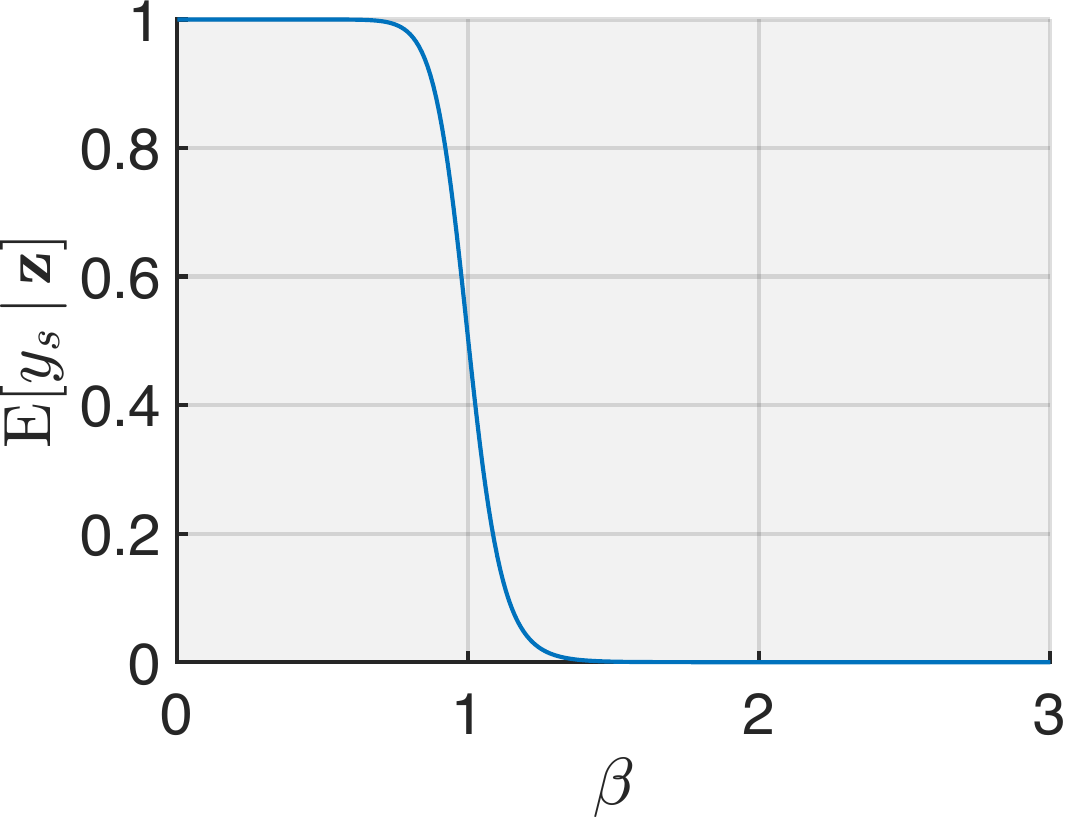}}
\vspace*{-1mm}
    \caption{Expected in-distribution label as a function of noise ratio $\beta = \sigma_{o}/\sigma_{s}$.
    \vspace*{-4mm}}
    \label{fig:beta_impact}
\end{figure}

Our model requires to specify the in-distribution variance $\sigma_s^2$ and the OOD variance $\sigma_o^2$. In this section, we present how these can be chosen. The OOD variance should be greater than the in-distribution, hence we can write $\sigma_o = \beta\sigma_s$, with $\beta = \sigma_o/\sigma_s > 1$. Hence, $\beta$ is a measure of noise ratio for the in-distribution and OOD samples.  Using this relationship, Eq. (\ref{eq:logit}) simplifies to
\begin{equation}
u \, = \, -\frac{1}{2\sigma_s^2}\left(\frac{\beta^2-1}{\beta^2}\right)\|\eeps\|^2 \, + \, D\log\beta
\end{equation}
For OOD samples, the expected value of ``logit'' $u$ is then given by
\begin{equation}
\begin{split}
\overline{u}_\beta &\, = \, \EX_{\eeps \sim \mathcal{N}(\bm{0}, \sigma_o^2 I)}\!\left[-\frac{1}{2\sigma_s^2}\left(\frac{\beta^2-1}{\beta^2}\right)\|\eeps\|^2 \, + \, D\log\beta\right] \\
&\, = \, -\frac{1}{2\sigma_s^2}\left(\frac{\beta^2-1}{\beta^2}\right)\underbrace{\EX_{\eeps \sim \mathcal{N}(\bm{0}, \sigma_o^2 I)}\!\left[\|\eeps\|^2\right]}_{\sigma_o^2\,=\,\beta^2\sigma_s^2} \, + \, D\log\beta \\[-5pt]
&\, = \, -\frac{1}{2}\left(\beta^2-1\right) \, + \, D\log\beta
\end{split}
\end{equation}
Figure \ref{fig:beta_impact} show how the expected in-distribution prediction $\mathbb{E}[y_s \, | \, \zz] = \mathrm{sigmoid}(\overline{u}_\beta)$ varies as function of $\beta$, for $D=16$ (the dimension used in our experiments). In this case, to have near-zero probability for OOD samples, one can choose any $\beta > 1.5$. In our experiments, we selected $\beta=2$.

\section{Experimental Settings}
\label{sec:experiments}

We evaluate NC-TTT on several TTT datasets, following the protocol of previous works. These benchmarks emulate different challenging domain shift scenarios, which help evaluating the effectiveness of our approach. As in \cite{ttt,clust3}, these benchmarks are categorized as \emph{common corruptions}, and \emph{sim-to-real} domain shift.

For \emph{common corruptions}, we evaluate our method on CIFAR-10-C and CIFAR-100-C \cite{cifar10c&10.1}. This family of domain shifts include 15 different corruptions such as Gaussian noise, JPEG compression, among others. Each corruption has 5 different levels of severity with 10,000 images, which amounts to 75 different testing scenarios. For each of the aforementioned datasets, CIFAR-10 and CIFAR-100 are used as source domains, with 10 and 100 classes respectively. Finally, the challenging large-scale VisDA-C \cite{visda} dataset corresponds to the \emph{sim-to-real domain shift}. The source domain comprises a training set of 152,397 images of 3D renderings from 12 different classes, while the test set consists in 72,372 video frames of the same categories. 

\begin{figure}[t!]
   \centering
   \includegraphics[width=1.02\linewidth]{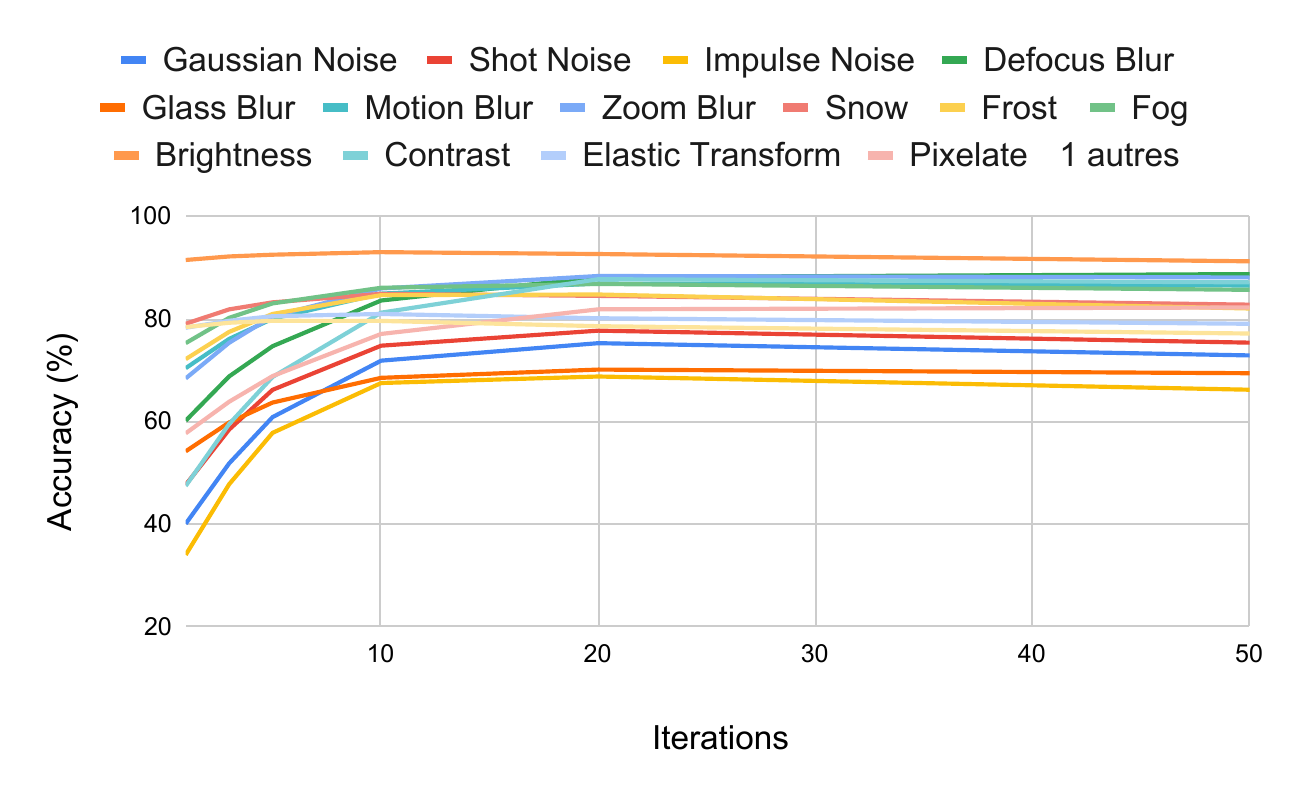} 
   \vspace*{-1mm}
   \caption{Evolution of accuracy on all corruptions in CIFAR-10-C.
   \vspace*{-4mm}}   
   \label{fig:cifar10cbar}
\end{figure}

\begin{table*}[t!]
    \centering
    \begin{small}
    \dorowcolors
    \resizebox{\textwidth}{!}{
    \begin{tabular}{l|cccccccc}
    \toprule
        ~ & ResNet50  & LAME~\cite{lame2022} & PTBN~\cite{PTBN}  & TENT~\cite{tent2021} & TTT~\cite{ttt} & TTT++~\cite{ttt++} & ClusT3~\cite{clust3} & NC-TTT (ours)  \\ \midrule
        Gaussian Noise & 21.01 & 22.90\ppm0.07 & 57.23\ppm0.13 & 57.15\ppm0.19 & 66.14\ppm0.12 & 75.87\ppm5.05 & \textbf{76.01\ppm0.19} & 75.30\ppm0.04 \\ 
        Shot noise & 25.77 & 27.11\ppm0.13 & 61.18\ppm0.03 & 61.08\ppm0.18 & 68.93\ppm0.06 & 77.18\ppm1.36 & 77.67\ppm0.17 & \textbf{77.74\ppm0.05}\\ 
        Impulse Noise & 14.02 & 30.99\ppm0.15 & 54.74\ppm0.13 & 54.63\ppm0.15 & 56.65\ppm0.03 & \textbf{70.47\ppm2.18}  & 69.76\ppm0.15 & 68.80\ppm0.11 \\ 
        Defocus blur & 51.59 & 45.16\ppm0.13 & 81.61\ppm0.07 & 81.39\ppm0.22 & 88.11\ppm0.08 & 86.02\ppm1.35 & 87.85\ppm0.11 & \textbf{88.77\ppm0.09} \\ 
        Glass blur & 47.96 & 36.58\ppm0.06 & 53.43\ppm0.11 & 53.36\ppm0.14 & 60.67\ppm0.06 & 69.98\ppm1.62 & \textbf{71.34\ppm0.15} & 70.15\ppm0.16 \\ 
        Motion blur & 62.30 & 55.41\ppm0.15 & 78.20\ppm0.28 & 78.04\ppm0.17 & 83.52\ppm0.03 & 85.93\ppm0.24 & 86.10\ppm0.11 & \textbf{86.93\ppm0.05}\\ 
        Zoom blur & 59.49 & 51.48\ppm0.20 & 80.29\ppm0.13 & 80.26\ppm0.22 & 87.25\ppm0.03 & \textbf{88.88\ppm0.95} & 86.68\ppm0.05 & 88.40\ppm0.06 \\ 
        Snow & 75.41 & 66.14\ppm0.12 & 71.59\ppm0.21 & 71.59\ppm0.04 & 79.29\ppm0.05 & 82.24\ppm1.69 & 83.71\ppm0.09 & \textbf{84.92\ppm0.08} \\ 
        Frost & 63.14 & 50.03\ppm0.22 & 68.77\ppm0.25 & 68.52\ppm0.20 & 79.84\ppm0.11 & 82.74\ppm1.63 & 83.69\ppm0.03 & \textbf{84.79\ppm0.05} \\ 
        Fog & 69.63 & 64.56\ppm0.19 & 75.79\ppm0.05 & 75.73\ppm0.10 & 84.46\ppm0.09 & 84.16\ppm0.28 & 85.12\ppm0.13 & \textbf{86.85\ppm0.10} \\ 
        Brightness & 90.53 & 84.27\ppm0.10 & 84.97\ppm0.05 & 84.77\ppm0.13 & 91.23\ppm0.08 & 89.07\ppm1.20 & 91.52\ppm0.02 & \textbf{93.05\ppm0.03} \\ 
        Contrast & 33.88 & 31.46\ppm0.23 & 80.81\ppm0.15 & 80.70\ppm0.15 & \textbf{88.58\ppm0.09} & 86.60\ppm1.39 & 84.40\ppm0.11 & 87.78\ppm0.15 \\ 
        Elastic transform & 74.51 & 64.23\ppm0.10 & 67.14\ppm0.17 & 67.13\ppm0.10 & 75.69\ppm0.10 & 78.46\ppm1.83 & \textbf{82.04\ppm0.17} & 80.99\ppm0.11 \\ 
        Pixelate & 44.43 & 39.32\ppm0.08 & 69.17\ppm0.31 & 68.70\ppm0.29 & 76.35\ppm0.19 & \textbf{82.53\ppm2.01} & 82.03\ppm0.09 & 82.26\ppm0.11 \\ 
        JPEG compression & 73.61 & 66.19\ppm0.02 & 65.86\ppm0.05 & 65.83\ppm0.07 & 73.10\ppm0.19 & 81.76\ppm1.58 & \textbf{83.24\ppm0.10} & 79.66\ppm0.06  \\ \midrule
        Average & 53.82 & 49.06 & 70.05 & 69.93 & 77.32 & 81.46 & 82.08 & \bf 82.43 
        \\
        \bottomrule 
    \end{tabular}}
    \end{small}
    \vspace*{-1mm}
    \caption{Accuracy (\%) on CIFAR-10-C dataset with Level 5 corruption for NC-TTT compared to previous TTA and TTT methods.}
	\label{tab:Cifar10cComparison}
\end{table*}

\begin{table*}[h!]
    \centering
    \begin{small}
    \dorowcolors
    \begin{tabular}{l|ccccccc}
    \toprule
        ~ & ResNet50  & LAME~\cite{lame2022} & PTBN~\cite{PTBN}  & TENT~\cite{tent2021} & TTT~\cite{ttt} & ClusT3~\cite{clust3} & NC-TTT (ours) \\ \midrule
        Gaussian Noise    & 12.67                        & 10.55\ppm0.08                     & 43.00\ppm0.16 & 43.17\ppm0.24 & 33.99\ppm0.11 & \textbf{49.77\ppm0.18} & 46.03\ppm0.12          \\
        Shot noise        & 14.79                        & 12.58\ppm0.04                     & 44.57\ppm0.16 & 44.47\ppm0.23 & 36.55\ppm0.08 & \textbf{50.54\ppm0.16} & 47.04\ppm0.14          \\
        Impulse Noise     & \phantom{0}6.47                         & 5.83\ppm0.07                      & 36.76\ppm0.11 & 36.64\ppm0.28 & 26.87\ppm0.08 & \textbf{44.35\ppm0.31} & 41.53\ppm0.11          \\
        Defocus blur      & 29.97                        & 29.07\ppm0.11                     & 66.68\ppm0.06 & 66.74\ppm0.06 & 65.96\ppm0.14 & 64.40\ppm0.12          & \textbf{67.00\ppm0.09}    \\
        Glass blur        & 21.36                        & 19.58\ppm0.02                     & 45.17\ppm0.08 & 45.09\ppm0.06 & 34.90\ppm0.01 & \textbf{50.78\ppm0.24} & 48.08\ppm0.07          \\
        Motion blur       & 39.60                         & 41.26\ppm0.09                     & 62.61\ppm0.17 & 62.54\ppm0.23 & 57.10\ppm0.10 & 62.62\ppm0.15          & \textbf{64.31\ppm0.02} \\
        Zoom blur         & 35.75                        & 34.93\ppm0.02                     & 65.36\ppm0.03 & 65.29\ppm0.05 & 62.90\ppm0.07 & 63.81\ppm0.08          & \textbf{66.24\ppm0.25} \\
        Snow              & 42.05                        & 43.58\ppm0.20                     & 52.82\ppm0.27 & 52.31\ppm0.16 & 54.97\ppm0.03 & 55.84\ppm0.12          & \textbf{58.70\ppm0.10}   \\
        Frost             & 31.44                        & 32.67\ppm0.12                     & 51.92\ppm0.09 & 51.79\ppm0.23 & 54.60\ppm0.16  & 55.46\ppm0.06          & \textbf{58.55\ppm0.11} \\
        Fog               & 30.96                        & 35.95\ppm0.12                     & 55.78\ppm0.05 & 55.91\ppm0.28 & 55.80\ppm0.09 & 51.39\ppm0.07          & \textbf{57.73\ppm0.17} \\
        Brightness        & 61.80                         & 64.84\ppm0.03                     & 66.20\ppm0.06 & 66.47\ppm0.06 & 73.25\ppm0.06 & 66.71\ppm0.11          & \textbf{71.36\ppm0.10}  \\
        Contrast          & 12.31                        & 15.50\ppm0.04                     & 60.84\ppm0.15 & 60.91\ppm0.19 & 60.97\ppm0.09 & 54.67\ppm0.05          & \textbf{61.53\ppm0.20}  \\
        Elastic transform & 53.06                        & 51.32\ppm0.13                     & 56.38\ppm0.04 & 56.43\ppm0.33 & 53.51\ppm0.04 & 59.44\ppm0.27          & \textbf{60.25\ppm0.04} \\
        Pixelate          & 26.08                        & 27.65\ppm0.02                     & 58.21\ppm0.14 & 58.19\ppm0.22 & 50.39\ppm0.05 & 60.75\ppm0.09          & \textbf{61.17\ppm0.33} \\
        JPEG compression  & 52.19                        & 49.95\ppm0.07                     & 51.65\ppm0.16 & 51.30\ppm0.16 & 49.62\ppm0.09 & \textbf{59.94\ppm0.12} & 55.69\ppm0.09          \\
        Average           & 31.37                        & 31.68 & 54.53 & 54.48 & 51.43 & 56.70 & \bf 57.68 
        \\
        \bottomrule 
    \end{tabular}
    \end{small}
    \vspace*{-1mm}
    \caption{Accuracy (\%) on CIFAR-100-C dataset with Level 5 corruption for NC-TTT and the works from the \emph{state-of-the-art}.
    \vspace*{-4mm}}
	\label{tab:Cifar100cComparison}
\end{table*}

\mypar{Source training.} The cross-entropy and auxiliary losses are jointly trained on the source dataset. We explored different architectural choices for each setting. For common corruptions (i.e. CIFAR-10/100-C), we define the projector as a $1\!\times\!1$ convolutional layer that reduces the number of channels to $D=96$ to later be flattened for classification. We utilize a discriminator composed of two linear layers with a Batch Norm layer and Leaky ReLU in between, and a hidden dimension of $1024$ in the intermediate layer. For this particular case, we use the tuple $(\sigma_{s}=0, \sigma_{o}=0.015)$, which was experimentally determined as it produced the best performance. The model is trained using 128 images per batch for 350 epochs using SGD, an initial learning rate of 0.1, and a multi-step scheduler with a decreasing factor of 10 at epochs 150 and 250. Due to the challenging nature of the \emph{sim-to-real} domain shift from VisDA-C, we escalate the architecture to make it able to learn more source domain information. We utilize a $1\!\times\!1$ convolutional projector with an output number of channels of $D=16$. As opposed to flattening the features, we also employ two $1\!\times\!1$ convolutional layers for the discriminator, with an intermediate number of channels of $1024$. The noise values are sampled with $(\sigma_{s}=0.025, \sigma_{o}=0.05)$ and added \emph{pixel-wise} to the projected feature maps. Following related works' protocol for VisDA-C, we use an ImageNet-pre-trained model \cite{imagenet} as a warm start, to then perform the source training with a batch size of 50 for 100 epochs with SGD and a learning rate of 0.01. ResNet50 \cite{resnet} is the chosen architecture for all datasets.

\mypar{Test-time adaptation.} Adaptation is performed on the encoder's blocks (including BatchNorm layers). If the auxiliary task is plugged to the third layer block, for instance, the weights of all the previous blocks will be optimized. The source training on CIFAR-10 is used to adapt for CIFAR-10-C. In an analog way, CIFAR-100 is utilized to adapt for CIFAR-100-C. For all this cases, the ADAM optimizer with a learning rate of $10^{-5}$ is used in batches of 128 images. As for VisDA-C, a batch size of 50 is employed with a learning rate of $10^{-4}$. The weights of the source model are restored after each batch.

\begin{figure*}[ht!]
    \centering
    \begin{small}\setlength{\tabcolsep}{2pt}
    \begin{tabular}{cc}    
\includegraphics[width=0.45\linewidth]{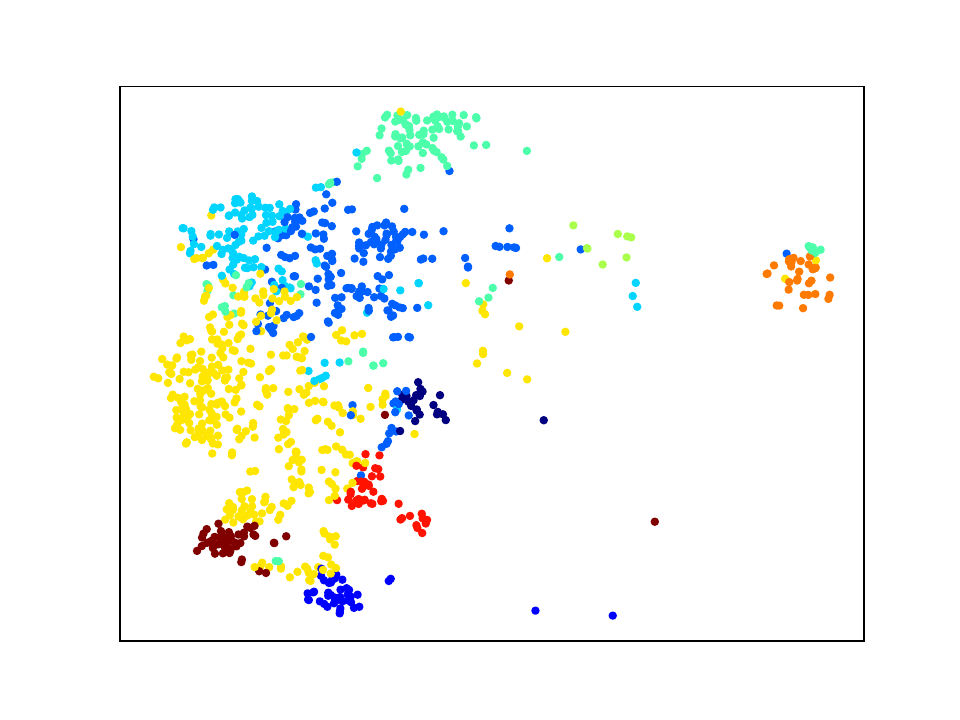} &     
    \includegraphics[width=0.45\linewidth]{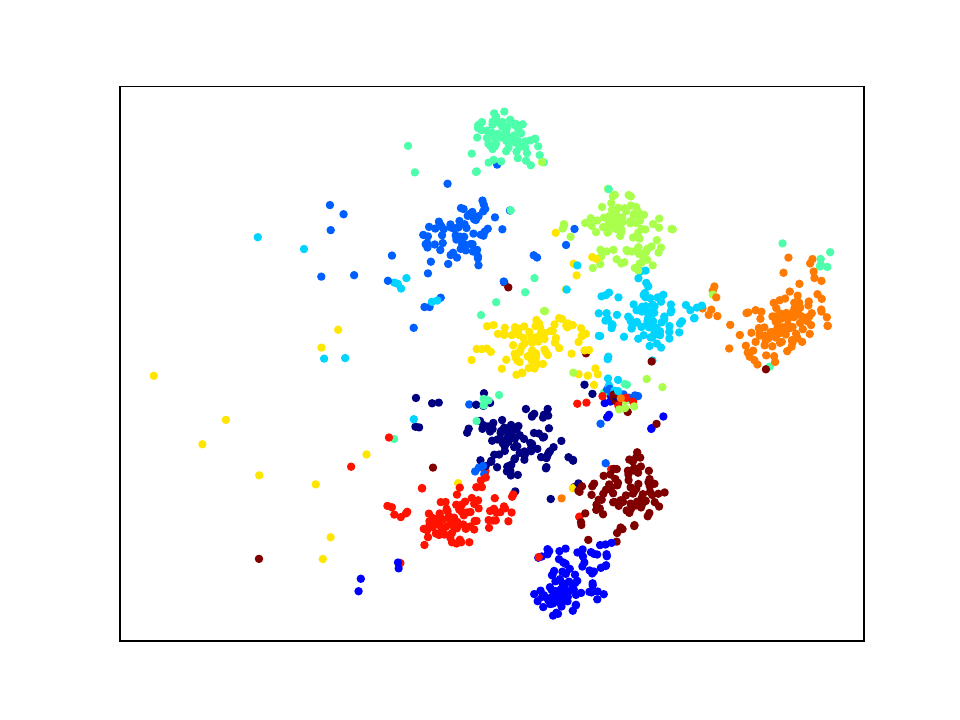}\\[-16pt]
    (a) Prediction (before adaptation) & (b) Prediction (after adaptation) \\
    \includegraphics[width=0.45\linewidth]{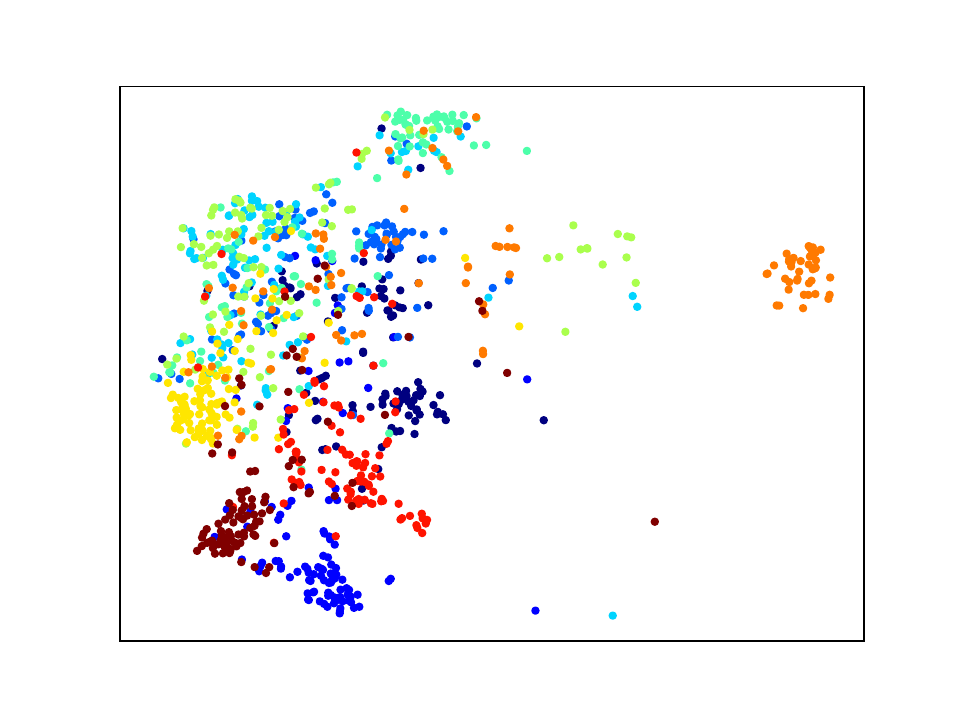} &
    \includegraphics[width=0.45\linewidth]{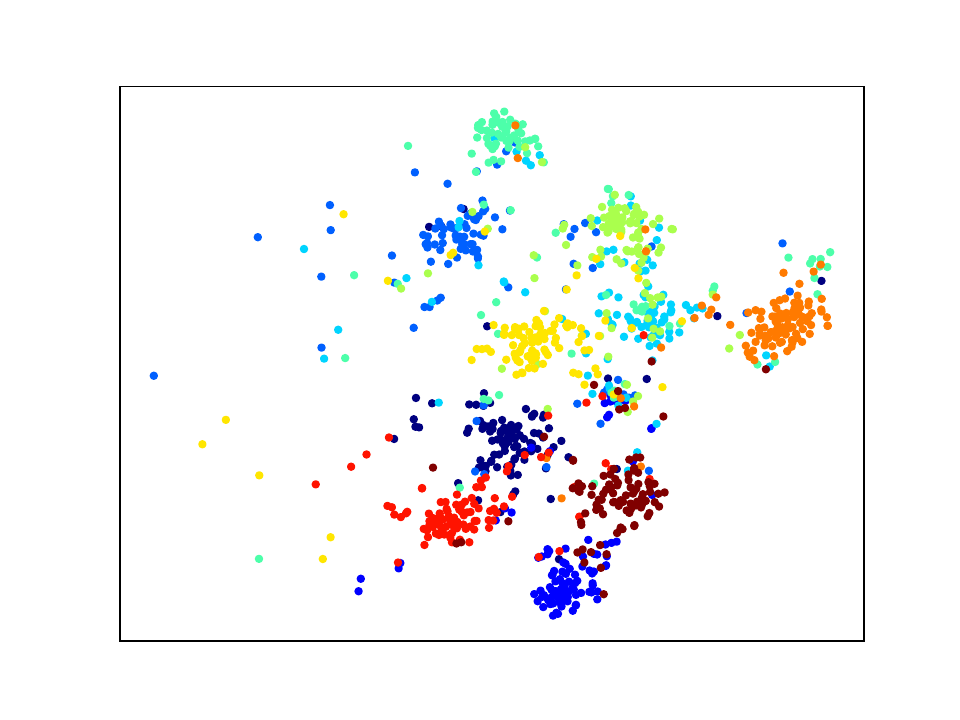}\\[-16pt]
    (c) Ground truth (before adaptation) & (d) Ground truth (after adaptation)\\[2pt]
    \end{tabular}
    \end{small}
    \vspace*{-1mm}
    \caption{t-SNE visualizations depict shot noise characteristics in the features extracted from NC-TTT. Panels (a) and (b) illustrate the model predictions without and with 20 iterations of adaptation, respectively. Panels (c) and (d) showcase the ground truth labels in the absence of adaptation and for the adapted representations, respectively.
    \vspace*{-4mm}}
    \label{fig:tSNEcifar10c}
\end{figure*}

\mypar{Benchmarking.} We compare the performance of NC-TTT with previous works from the \emph{state-of-the-art} in TTT and TTA. Chosen works in TTA include PTBN~\cite{PTBN}, TENT~\cite{tent2021}, and LAME~\cite{lame2022}, whereas for TTT we consider TTT~\cite{ttt}, TTT++~\cite{ttt++}, and ClusT3~\cite{clust3}. We utilize the source model (named ResNet50 in our results) without adaptation to measure accuracy gains.

\section{Results}
\label{sec:results}

In this section, we present the experimental results obtained from NC-TTT and compare them against the \emph{state-of-the-art}. In accordance with previous TTT research, we also offer insights on the working mechanisms that take part in the success of our technique.

\subsection{Image classification on common corruptions}

We assess the performance of ClusT3 using the CIFAR-10/100-C dataset, considering 15 distinct corruptions. Subsequently, our experiments concentrate exclusively on Level 5, recognized as the most demanding adaptation scenario. Comprehensive results for all severity levels are provided in the Supplementary material.

The data presented in Fig~\ref{fig:cifar10cbar} reveals that peak accuracy is typically reached around 20 iterations, depending on the specific corruption type. Remarkably, accuracy remains stable even beyond the 20th iteration. In the case of certain corruptions, specifically the ones with noise such as Impulse Noise which significantly degrade the image quality, we observe a decline in performance with an increase in the number of adaptation iterations.

As shown in Table~\ref{tab:Cifar10cComparison}, NC-TTT achieves an average improvement of 30.61$\%$ with respect to the baseline (i.e. ResNet50), and obtains a considerable advantage in all the different corruptions. Moreover, our method achieves to outperform ClusT3 in most corruptions and in average for the whole dataset. It is worth noticing that, besides the strong relation of NC-TTT to Gaussian-like noise, the performance on the \emph{Gaussian Noise} corruption is not necessarily the highest, which could be due to the fact that the auxiliary task does not bias the model towards any type of domain shift. Table~\ref{tab:Cifar100cComparison} shows a more surprising trend on CIFAR-100-C, as our technique outperforms the closest competitor on the majority of the corruptions, and obtains an average improvement of $26.31\%$ with respect to ResNet50. Based on the above, NC-TTT can approximate the source information even when: a) the number of classes increases, and b) the auxiliary task works at a smaller scale as the main classification task.

Figure~\ref{fig:tSNEcifar10c} demonstrates the impact of NC-TTT during adaptation through t-SNE plots showcasing the target feature maps before and after adaptation, along with the associated model predictions. The challenging corruption of shot noise becomes more manageable with the assistance of NCE, contributing to improved predictions by refining the clustering of diverse class samples within the target dataset.


\subsection{Image classification on sim-to-real domain shift}

For adaptation on VisDA-C, the first encoder's layer block is chosen for the auxiliary task. The obtained results concur with previous works \cite{ttt,clust3}, in that the first layers of the network's encoder are sufficient for adaptation.  

As shown in Table~\ref{tab:visda}, NC-TTT obtains a competitive performance with respect to previous works on VisDA-C. The severe domain shift in this dataset makes it a very challenging scenario, as can be seen when testing the source model. NC-TTT obtains a gain of $16.19\%$ in accuracy, and surpasses previous methods by an important margin.

\begin{table}[ht!]
\centering
\begin{small}
\dorowcolors
\begin{tabular}{lc}
\toprule
\textbf{Method}        & \textbf{Acc. (\%)}   \\ \midrule
ResNet50      & 46.31              \\
LAME-L~\cite{lame2022}        & 22.02\ppm 0.23 \\
LAME-K~\cite{lame2022}        & 42.89\ppm 0.14 \\
LAME-R~\cite{lame2022}        & 19.33\ppm 0.11 \\
PTBN~\cite{PTBN}          & 60.33\ppm 0.04 \\
TENT~\cite{tent2021}          & 60.34\ppm 0.05 \\
TTT~\cite{ttt}           &  40.57\ppm 0.02                   \\
ClusT3~\cite{clust3}         & 61.91\ppm 0.02             \\ 
NC-TTT (ours) & 
\bf 62.71\ppm 0.09 
\\
\bottomrule
\end{tabular}
\end{small}
\vspace*{-1mm}
\caption{Results on VisDA-C.
\vspace*{-4mm}}
\label{tab:visda}
\end{table}

\section{Conclusions}
\label{sec:conclusions}

We proposed NC-TTT, a Test-Time Training method based on the popular theory of Noise-Contrastive Estimation. Our method learns a proximal representation of the source domain by discriminating between noisy views of feature maps. The entire model can be added on top of any given layer of a CNN's encoder, and comprises only a linear projector and a classifier.

The proposed experiments support already established hypothesis of TTT, which states that adaptation in the first encoder's layer blocks (e.g. first or second) is often sufficient to recover the model's performance on a new domain. NC-TTT is evaluated on different challenging benchmarks, and its performance is compared against recent \emph{state-of-the-art} methods in the field. 

This work leads to interesting questions that can be addressed as future work. First, different types of added noise could be explored to analyze their impact in the learning of the auxiliary task. A similar framework can eventually be derived for different distributions. Moreover, and as an open question partaking all the existent TTT methods, the exact mechanisms that allow auxiliary tasks to learn domain-related information are unclear. This is especially intriguing considering that the scale of such tasks is small compared to the classification task. Their properties and their relation with the models' performance a suitable research direction.

\newpage

\section*{\centering NC-TTT: A Noise Contrastive Approach for Test-Time Training -- Supplementary Material}

\appendix
\section{Deriving the posterior of Equation (6)}\label{sec:proof}
We start with the definition of the posterior in Eq. (5):
\begin{equation*}
\begin{split}
& p(y_s=1 \, | \, \wt{\zz}_{i,m}) \, = \, \\[-6pt]
& \ \qquad
\frac{
\sigma_s^{-D} \exp\!\left(\!-\frac{1}{2\sigma_s^2}\|\eeps\|^2\right)
}{
\sigma_s^{-D} \exp\!\left(\!-\frac{1}{2\sigma_s^2}\|\eeps\|^2\right) + \sigma_o^{-D} \exp\!\left(\!-\frac{1}{2\sigma_o^2}\|\eeps\|^2\right)
},
\end{split}
\end{equation*}
where $\eeps = \wt{\zz}_i-\zz_i$. Multiplying the numerator and denominator by $\sigma_s^{D} \exp\!\big(\frac{1}{2\sigma_s^2}\|\eeps\|^2\big)$
gives 
\begin{equation*}
\begin{split}
& p(y_s=1 \, | \, \wt{\zz}_{i,m}) \, = \, \\[-6pt]
& \ \qquad
\frac{
 1
}{
1 \, + \, \big(\sigma_s/\sigma_o\big)^{D} \exp\!\left(\!-\!\frac{1}{2}(1/\sigma_s^2 - 1/\sigma_o^2)\|\eeps\|^2\right)
}
\end{split}
\end{equation*}
We then use the following equality
\begin{equation*}
\begin{split}
\big(\sigma_s/\sigma_o\big)^{D} & \, = \, \exp\left(\log\big(\sigma_s/\sigma_o\big)^{D}\right) \\
& \, = \, \exp\left(D\log\big(\sigma_s/\sigma_o\big)\right)
\end{split}
\end{equation*}
to obtain 
\begin{equation*}
\begin{split}
& p(y_s=1 \, | \, \wt{\zz}_{i,m}) \, = \, \\[-6pt]
& \ \qquad
\frac{
 1
}{
1 \, + \, \exp\!\left(\!-\!\frac{1}{2}(1/\sigma_s^2 - 1/\sigma_o^2)\|\eeps\|^2 \, + \, D\log\big(\sigma_s/\sigma_o\big)\right)
}
\end{split}
\end{equation*}
Last, since $\log\big(\sigma_s/\sigma_o\big) = -\log\big(\sigma_o/\sigma_s\big)$, we finally get $p(y_s=1 \, | \, \wt{\zz}) = 1/\big(1+ \exp(-u)\big)$ with
\begin{equation*}
u \, = \, \frac{1}{2}\left(\frac{1}{\sigma_o^2} - \frac{1}{\sigma_s^2} \right) \|\eeps\|^2 \, + \, D\log\left(\frac{\sigma_o}{\sigma_s}\right) \qed
\end{equation*}

\section{Results on different levels of CIFAR-10-C corruptions}

We evaluate NC-TTT on the remaining severity levels of CIFAR-10-C (see Tables \ref{tab:Cifar10cComparisonL4}-\ref{tab:Cifar10cComparisonL1}). Accuracy decreases on all methods as the severity augments, but NC-TTT outperforms the closest competitors from the \emph{state-of-the-art} on the majority of the corruptions and in average for the whole dataset. To be more precise, when considering NC-TTT at a lower severity level (refer to Table~\ref{tab:Cifar10cComparisonL1}), it demonstrates superior performance across all corruptions, with the exception of \emph{Gaussian Noise} and \emph{JPEG compression}. This is noteworthy, given that Gaussian noise is introduced to features during training.

\begin{table*}[!t]
    \centering
    \begin{small}
    \dorowcolors
    \resizebox{\textwidth}{!}{
    \begin{tabular}{l|llllllll}
    \toprule
        ~ & ResNet50  & LAME & PTBN  & TENT & TTT & TTT++ & ClusT3 & NC-TTT (ours) \\ \midrule
        Gaussian Noise & 28.02 & 26.22\ppm0.21 & 61.39\ppm0.10 & 61.19\ppm0.26 & 70.63\ppm0.04 & 78.70\ppm4.28 & \textbf{79.14\ppm0.03} &  78.09\ppm0.02\\ 
        Shot noise & 38.33 & 37.06\ppm0.17 & 66.57\ppm0.06 & 66.2\ppm0.18 & 75.18\ppm0.04 & 80.12\ppm0.12 & \textbf{81.51\ppm0.15} & 81.33\ppm0.10\\ 
        Impulse Noise & 46.12 & 45.03\ppm0.17 & 63.56\ppm0.20 & 62.98\ppm0.19 & 65.91\ppm0.04 & 70.64\ppm0.53 & \textbf{76.95\ppm0.07} & 75.52\ppm0.08\\ 
        Defocus blur & 67.33 & 67.70\ppm0.07 & 85.48\ppm0.12 & 85.32\ppm0.18 & \textbf{91.95\ppm0.02} & 81.75\ppm0.43 & 90.33\ppm0.09 & 91.91\ppm0.05\\ 
        Glass blur & 34.42 & 32.63\ppm0.09 & 52.26\ppm0.04 & 52.08\ppm0.15 & 60.44\ppm0.05 & 62.85\ppm0.50 & \textbf{71.09\ppm0.17} & 69.95\ppm0.09\\ 
        Motion blur & 63.71 & 64.00\ppm0.01 & 80.78\ppm0.12 & 80.75\ppm0.09 & 86.29\ppm0.10 & 68.42\ppm1.08 & 87.87\ppm0.11 & \textbf{89.02\ppm0.10}\\ 
        Zoom blur & 61.27 & 62.12\ppm0.21 & 83.33\ppm0.11 & 83.28\ppm0.10 & 89.90\ppm0.04 & 70.74\ppm2.05 & 88.86\ppm0.04 & \textbf{90.96\ppm0.05}\\ 
        Snow & 72.15 & 72.18\ppm0.04 & 73.25\ppm0.16 & 73.17\ppm0.25 & 81.25\ppm0.02 & 52.43\ppm0.56 & 84.30\ppm0.07 & \textbf{85.36\ppm0.06}\\ 
        Frost & 62.27 & 61.72\ppm0.06 & 73.41\ppm0.22 & 73.54\ppm0.16 & 83.83\ppm0.04 & 52.80\ppm2.67 & 87.17\ppm0.07 & \textbf{88.08\ppm0.02}\\ 
        Fog & 81.86 & 82.07\ppm0.03 & 83.88\ppm0.06 & 83.81\ppm0.09 & 90.62\ppm0.05 & 41.75\ppm0.09 & 90.03\ppm0.02 & \textbf{92.07\ppm0.07}\\ 
        Brightness & 87.58 & 87.64\ppm0.08 & 86.81\ppm0.05 & 86.81\ppm0.23 & 92.87\ppm0.09 & 50.95\ppm2.19 & \textbf{92.99\ppm0.06} & \textbf{94.41\ppm0.03}\\ 
        Contrast & 68.62 & 69.02\ppm0.11 & 84.16\ppm0.09 & 84.23\ppm0.29 & 90.94\ppm0.07 & 45.28\ppm0.55 & \textbf{89.24\ppm0.07} & \textbf{91.52\ppm0.03}\\ 
        Elastic transform & 67.84 & 68.32\ppm0.03 & 76.44\ppm0.18 & 76.21\ppm0.08 & 84.03\ppm0.11 & 35.53\ppm1.51 & \textbf{86.74\ppm0.04} &86.39\ppm0.04 \\ 
        Pixelate & 56.3 & 55.94\ppm0.08 & 76.34\ppm0.10 & 76.40\ppm0.16 & 84.92\ppm0.15 & 33.64\ppm0.83 & 87.93\ppm0.03 & \textbf{88.06\ppm0.20}\\ 
        JPEG compression & 70.62 & 70.44\ppm0.18 & 69.64\ppm0.03 & 69.54\ppm0.05 & 76.46\ppm0.04 & 28.01\ppm1.75 & \textbf{85.11\ppm0.06} & 82.73\ppm0.07\\ \midrule
        Average & 60.43 & 60.14 & 74.48 & 74.37 & 81.68 & 56.91 & 85.28 & \textbf{85.69}\\ \bottomrule
    \end{tabular}}
    \end{small}
    \caption{Accuracy (\%) on CIFAR-10-C dataset with Level 4 corruption for NC-TTT compared to \emph{state-of-the-art}.}
	\label{tab:Cifar10cComparisonL4}
\end{table*}

\begin{table*}[!t]
    \centering
    \begin{small}
    \dorowcolors
    \resizebox{\textwidth}{!}{
    \begin{tabular}{l|llllllll}
    \toprule
        ~ & ResNet50  & LAME & PTBN & TENT & TTT & TTT++ & ClusT3 & NC-TTT (ours) \\ \midrule
        Gaussian Noise & 33.99 & 32.37\ppm0.16 & 64.55\ppm0.13 & 64.67\ppm0.17 & 74.10\ppm0.09 & 80.29\ppm0.81 & \textbf{81.55\ppm0.09} & 81.09\ppm0.09\\ 
        Shot noise & 46.35 & 45.83\ppm0.14 & 69.82\ppm0.08 & 70.04\ppm0.14 & 78.43\ppm0.07 & 82.46\ppm0.37 & \textbf{84.12\ppm0.02} & 84.03\ppm0.09\\ 
        Impulse Noise & 59.90 & 59.43\ppm0.13 & 72.08\ppm0.14 & 71.95\ppm0.33 & 76.32\ppm0.10 & 79.20\ppm0.38 & \textbf{83.75\ppm0.01} & 83.73\ppm0.05\\ 
        Defocus blur & 79.29 & 79.67\ppm0.11 & 87.62\ppm0.17 & 87.39\ppm0.05 & 93.25\ppm0.06 & 87.68\ppm0.38 & 91.74\ppm0.07 & \textbf{93.51\ppm0.08}\\ 
        Glass blur & 47.29 & 46.36\ppm0.10 & 63.29\ppm0.11 & 63.26\ppm0.21 & 72.09\ppm0.11 & 72.52\ppm0.56 & \textbf{79.78\ppm0.02} & 79.25\ppm0.08\\ 
        Motion blur & 63.42 & 63.72\ppm0.07 & 81.13\ppm0.13 & 80.99\ppm0.08 & 86.48\ppm0.09 & 69.59\ppm1.38 & 88.02\ppm0.10 & \textbf{89.02\ppm0.02}\\ 
        Zoom blur & 67.86 & 68.23\ppm0.08 & 84.57\ppm0.11 & 84.34\ppm0.06 & 91.00\ppm0.02 & 73.23\ppm2.33 & 89.90\ppm0.07 & \textbf{91.86\ppm0.10}\\ 
        Snow & 74.93 & 74.78\ppm0.05 & 75.08\ppm0.14 & 75.14\ppm0.19 & 83.90\ppm0.07 & 57.96\ppm1.02 & 86.22\ppm0.07 & \textbf{87.17\ppm0.05}\\ 
        Frost & 64.54 & 64.16\ppm0.08 & 74.15\ppm0.04 & 73.98\ppm0.14 & 84.13\ppm0.10 & 49.94\ppm3.53 & \textbf{87.37\ppm0.07} & 88.05\ppm0.05\\ 
        Fog & 85.73 & 85.98\ppm0.16 & 86.57\ppm0.09 & 86.38\ppm0.15 & 92.19\ppm0.08 & 52.89\ppm4.13 & 91.83\ppm0.01 & \textbf{93.55\ppm0.01}\\ 
        Brightness & 88.93 & 88.67\ppm0.08 & 87.50\ppm0.19 & 87.44\ppm0.01 & 93.53\ppm0.09 & 57.96\ppm1.32 & 93.31\ppm0.04 & \textbf{94.61\ppm0.01}\\ 
        Contrast & 79.66 & 79.99\ppm0.05 & 85.63\ppm0.05 & 85.46\ppm0.08 & 91.85\ppm0.09 & 53.44\ppm2.37 & 90.83\ppm0.05 & \textbf{92.54\ppm0.09}\\ 
        Elastic transform & 75.67 & 75.96\ppm0.14 & 82.72\ppm0.14 & 82.56\ppm0.15 & 90.09\ppm0.10 & 36.49\ppm3.72 & 89.33\ppm0.11 & \textbf{90.95\ppm0.03}\\ 
        Pixelate & 74.83 & 75.12\ppm0.04 & 82.17\ppm0.14 & 81.91\ppm0.13 & 89.30\ppm0.10 & 33.41\ppm3.02 & 90.23\ppm0.06 & \textbf{91.44\ppm0.05}\\ 
        JPEG compression & 73.70 & 73.66\ppm0.16 & 71.54\ppm0.09 & 71.54\ppm0.15 & 78.95\ppm0.09 & 28.82\ppm2.74 & \textbf{86.55\ppm0.06} & 85.10\ppm0.01\\ \midrule
        Average & 67.74 & 67.60 & 77.89 & 77.80 & 85.04 & 61.06 & 87.64 & \textbf{88.39}\\ \bottomrule
    \end{tabular}}
    \end{small}
    \caption{Accuracy (\%) on CIFAR-10-C dataset with Level 3 corruption for NC-TTT compared to \emph{state-of-the-art}.}
	\label{tab:Cifar10cComparisonL3}
\end{table*}

\begin{table*}[!t]
    \centering
    \begin{small}
    \dorowcolors
    \resizebox{\textwidth}{!}{
    \begin{tabular}{l|llllllll}
    \toprule
         & ResNet50  & LAME & PTBN  & TENT & TTT & TTT++ & ClusT3 & NC-TTT (ours) \\ \midrule
        Gaussian Noise & 50.53 & 50.02\ppm0.24 & 71.31\ppm0.16 & 71.43\ppm0.08 & 81.18\ppm0.11 & 85.41\ppm2.26 & \textbf{86.07\ppm0.08} & 85.37\ppm0.08\\ 
        Shot noise & 69.27 & 69.47\ppm0.22 & 78.97\ppm0.19 & 79.02\ppm0.17 & 87.54\ppm0.10 & 88.79\ppm0.44 & 89.77\ppm0.04 & \textbf{90.15\ppm0.09}\\ 
        Impulse Noise & 68.57 & 68.68\ppm0.08 & 77.09\ppm0.13 & 77.03\ppm0.15 & 82.20\ppm0.13 & 84.27\ppm0.29 & 86.60\ppm0.03 & \textbf{86.89\ppm0.11}\\ 
        Defocus blur & 87.45 & 87.46\ppm0.14 & 88.20\ppm0.11 & 88.06\ppm0.06 & 93.67\ppm0.06 & 90.85\ppm0.42 & 92.87\ppm0.01 & \textbf{94.32\ppm0.01}\\ 
        Glass blur & 43.26 & 42.04\ppm0.19 & 62.66\ppm0.09 & 62.55\ppm0.11 & 71.33\ppm0.04 & 71.60\ppm1.95 & 78.81\ppm0.11 & \textbf{79.86\ppm0.06}\\ 
        Motion blur & 72.98 & 73.14\ppm0.06 & 83.51\ppm0.16 & 83.46\ppm0.10 & 89.57\ppm0.07 & 77.38\ppm1.12 & 89.78\ppm0.13 & \textbf{91.23\ppm0.04}\\ 
        Zoom blur & 74.89 & 75.23\ppm0.18 & 85.81\ppm0.21 & 85.79\ppm0.05 & 92.05\ppm0.10 & 80.30\ppm1.45 & 90.82\ppm0.04 & \textbf{92.76\ppm0.10}\\ 
        Snow & 71.11 & 70.78\ppm0.12 & 74.73\ppm0.11 & 74.69\ppm0.22 & 82.96\ppm0.08 & 68.56\ppm1.36 & 86.30\ppm0.04 & \textbf{87.55\ppm0.05}\\ 
        Frost & 76.67 & 76.46\ppm0.02 & 79.54\ppm0.15 & 79.41\ppm0.27 & 87.67\ppm0.03 & 63.66\ppm3.39 & 90.27\ppm0.10 & \textbf{91.09\ppm0.03}\\ 
        Fog & 88.51 & 88.55\ppm0.08 & 87.62\ppm0.10 & 87.60\ppm0.17 & 93.23\ppm0.04 & 64.26\ppm3.37 & 93.07\ppm0.04 & \textbf{94.46\ppm0.02}\\ 
        Brightness & 89.75 & 89.52\ppm0.01 & 88.09\ppm0.03 & 87.97\ppm0.14 & 93.69\ppm0.08 & 67.19\ppm1.23 & 93.64\ppm0.01 & \textbf{94.97\ppm0.04}\\ 
        Contrast & 84.58 & 84.87\ppm0.07 & 86.19\ppm0.17 & 86.41\ppm0.04 & 92.50\ppm0.12 & 62.90\ppm1.93 & 92.00\ppm0.01 & \textbf{3.50\ppm0.08}\\ 
        Elastic transform & 82.10 & 82.17\ppm0.10 & 83.69\ppm0.13 & 83.68\ppm0.08 & 90.98\ppm0.12 & 50.06\ppm2.37 & 90.37\ppm0.01 & \textbf{91.60\ppm0.05}\\ 
        Pixelate & 81.04 & 80.96\ppm0.13 & 82.92\ppm0.14 & 83.01\ppm0.07 & 90.61\ppm0.15 & 43.33\ppm3.31 & 91.28\ppm0.09 & \textbf{92.46\ppm0.06}\\ 
        JPEG compression & 76.06 & 75.92\ppm0.09 & 73.63\ppm0.02 & 73.56\ppm0.13 & 81.37\ppm0.11 & 28.26\ppm2.78 & \textbf{87.86\ppm0.08} & 86.27\ppm0.03\\ \midrule
        Average & 74.45 & 74.35 & 80.26 & 80.24 & 87.37 & 68.45 & 89.30 & \textbf{90.17}\\ \bottomrule
    \end{tabular}}
    \end{small}
    \caption{Accuracy (\%) on CIFAR-10-C dataset with Level 2 corruption for NC-TTT compared to \emph{state-of-the-art}.}
	\label{tab:Cifar10cComparisonL2}
\end{table*}

\begin{table*}[!t]
    \centering
    \begin{small}
    \dorowcolors
    \resizebox{\textwidth}{!}{
    \begin{tabular}{l|llllllll}
    \toprule
         & ResNet50  & LAME & PTBN  & TENT & TTT & TTT++ & ClusT3 & NC-TTT (ours) \\ \midrule
        Gaussian Noise & 71.38 & 71.35\ppm0.05 & 79.22\ppm0.13 & 79.52\ppm0.12 & 88.38\ppm0.12 & 90.14\ppm1.05 & \textbf{90.35\ppm0.05} & 90.29\ppm0.01\\ 
        Shot noise & 80.39 & 80.32\ppm0.07 & 82.21\ppm0.05 & 82.18\ppm0.15 & 90.43\ppm0.02 & 90.89\ppm0.29 & 91.42\ppm0.02 & \textbf{92.25\ppm0.02}\\ 
        Impulse Noise & 80.04 & 79.98\ppm0.09 & 82.39\ppm0.08 & 82.48\ppm0.15 & 88.23\ppm0.02 & 87.76\ppm0.06 & 90.51\ppm0.06 & \textbf{91.07\ppm0.05}\\ 
        Defocus blur & 90.17 & 89.9\ppm0.06 & 88.28\ppm0.04 & 88.26\ppm0.15 & 93.89\ppm0.04 & 91.51\ppm0.48 & 93.72\ppm0.09 & \textbf{95.12\ppm0.02}\\ 
        Glass blur & 40.96 & 39.87\ppm0.16 & 63.19\ppm0.05 & 63.22\ppm0.15 & 71.12\ppm0.07 & 72.12\ppm2.13 & 79.01\ppm0.21 & \textbf{79.78\ppm0.05}\\ 
        Motion blur & 82.78 & 82.81\ppm0.11 & 85.99\ppm0.09 & 85.89\ppm0.08 & 91.97\ppm0.05 & 84.11\ppm0.91 & 91.50\ppm0.13 & \textbf{93.15\ppm0.07}\\ 
        Zoom blur & 78.58 & 79.03\ppm0.06 & 86.19\ppm0.06 & 86.23\ppm0.04 & 92.21\ppm0.08 & 81.76\ppm1.38 & 90.87\ppm0.04 & \textbf{92.60\ppm0.07}\\ 
        Snow & 83.45 & 83.32\ppm0.11 & 82.94\ppm0.13 & 82.84\ppm0.35 & 88.90\ppm0.04 & 75.89\ppm0.75 & 90.33\ppm0.02 & \textbf{91.57\ppm0.03}\\ 
        Frost & 84.84 & 84.44\ppm0.10 & 83.88\ppm0.15 & 83.71\ppm0.24 & 91.17\ppm0.03 & 71.54\ppm3.13 & 92.19\ppm0.06 & \textbf{93.16\ppm0.08}\\ 
        Fog & 90.15 & 90.05\ppm0.05 & 88.31\ppm0.13 & 88.05\ppm0.06 & 93.71\ppm0.09 & 70.58\ppm1.29 & 93.64\ppm0.01 & \textbf{95.11\ppm0.03}\\ 
        Brightness & 90.35 & 90.24\ppm0.06 & 88.28\ppm0.09 & 88.35\ppm0.25 & 93.90\ppm0.06 & 64.40\ppm2.69 & 93.83\ppm0.05 & \textbf{95.28\ppm0.02}\\ 
        Contrast & 89.52 & 89.57\ppm0.07 & 87.98\ppm0.09 & 87.93\ppm0.08 & 93.61\ppm0.05 & 53.60\ppm3.80 & 93.61\ppm0.03 & \textbf{94.95\ppm0.06}\\ 
        Elastic transform & 82.46 & 82.72\ppm0.06 & 83.29\ppm0.17 & 83.28\ppm0.27 & 90.55\ppm0.09 & 39.92\ppm1.52 & 90.33\ppm0.06 & \textbf{91.62\ppm0.07}\\ 
        Pixelate & 87.27 & 87.18\ppm0.08 & 85.79\ppm0.12 & 85.81\ppm0.17 & 92.24\ppm0.01 & 36.04\ppm3.47 & 92.74\ppm0.04 & \textbf{93.84\ppm0.03}\\ 
        JPEG compression & 82.03 & 81.66\ppm0.07 & 79.72\ppm0.10 & 79.82\ppm0.14 & 86.86\ppm0.08 & 30.90\ppm1.18 & \textbf{90.90\ppm0.01} & 90.18\ppm0.05\\ \midrule
        Average & 80.96 & 80.83 & 83.17 & 83.17 & 89.81 & 69.41 & 91.00 & \textbf{92.00}\\ \bottomrule
    \end{tabular}}
    \end{small}
    \caption{Accuracy (\%) on CIFAR-10-C dataset with Level 1 corruption for NC-TTT compared to \emph{state-of-the-art}.}
	\label{tab:Cifar10cComparisonL1}
\end{table*}

\section{Hyperparameter search on VisDA-C}

\begin{figure}
    \centering
    \includegraphics[scale=0.32]{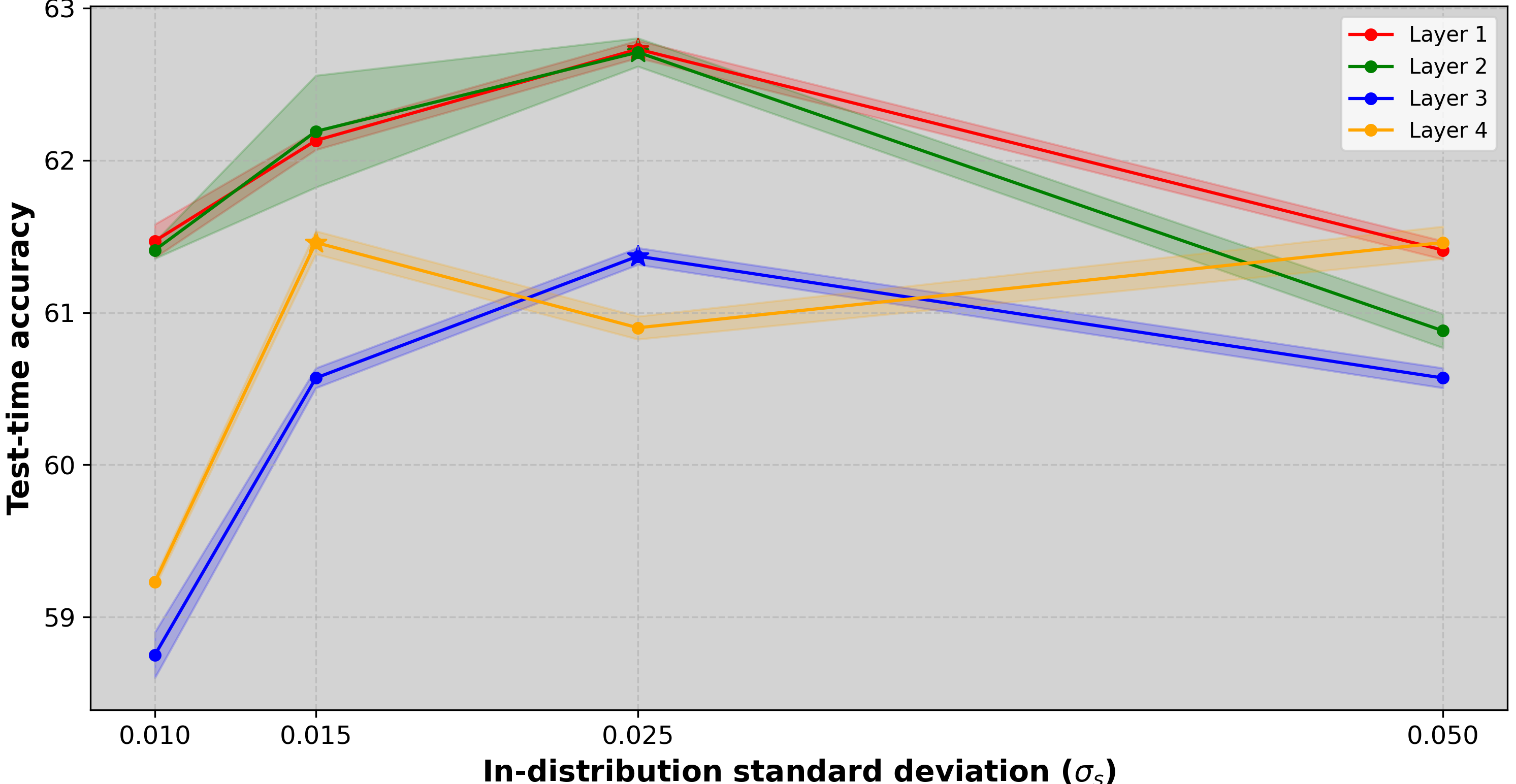}
    \caption{Test-time accuracy on different layer blocks with different in-distribution standard deviation.}
    \label{fig:hypervisda}
\end{figure}

In order to choose the best configuration for VisDA-C, we performed a hyperparameter search considering the four different layer blocks from ResNet50 as well as four different in-distribution noise standard deviation values (i.e., 0.01, 0.015, 0.025, 0.05). The results are obtained across three executions per combination. We show in Fig.\ref{fig:hypervisda} that the best performance can be generally obtained on the first layer of the network, consistent with previous results in the field \cite{ttt,ttt++,tttflow,clust3}. Furthermore, an in-distribution standard deviation $\sigma_{s} = 0.025$ is found to perform the best across all the different layers. 

\bibliographystyle{unsrtnat}
\bibliography{main} 

\end{document}